\documentclass[conference]{IEEEtran}
\IEEEoverridecommandlockouts
\usepackage{cite}
\usepackage{amsmath,amssymb,amsfonts}
\usepackage{algorithmic}
\usepackage{graphicx}
\usepackage{textcomp}
\usepackage{xcolor}
\usepackage{multirow}
\usepackage{arydshln}
\def\BibTeX{{\rm B\kern-.05em{\sc i\kern-.025em b}\kern-.08em
    T\kern-.1667em\lower.7ex\hbox{E}\kern-.125emX}}
   
\begin{document}

\title{A Blended Deep Learning Approach for Predicting User Intended Actions}

\author{
  \IEEEauthorblockN{
    Fei Tan\IEEEauthorrefmark{2}, 
    Zhi Wei\IEEEauthorrefmark{2}$^\ast$\thanks{$^\ast$Corresponding author.}, 
    Jun He\IEEEauthorrefmark{3}, 
    Xiang Wu\IEEEauthorrefmark{3},
    Bo Peng\IEEEauthorrefmark{3},
    Haoran Liu\IEEEauthorrefmark{2},
    and 
    Zhenyu Yan\IEEEauthorrefmark{3}$^\ast$}
  \IEEEauthorblockA{
    \IEEEauthorrefmark{2}Department of Computer Science,
      New Jersey Institute of Technology}
    \IEEEauthorblockA{\IEEEauthorrefmark{3}Digital Marketing,  Adobe Systems Inc.\\
       \{ft54, zhiwei, hl425\}@njit.edu, \{juhe, xianwu, bpeng, wyan\}@adobe.com}
      }

\maketitle

\begin{abstract}
User intended actions are widely seen in many areas. 
Forecasting these actions and taking proactive measures to optimize business outcome
is a crucial step towards sustaining the steady business growth.
In this work, we focus on predicting attrition, which is one of typical user intended actions. 
Conventional attrition predictive modeling strategies suffer a few inherent drawbacks. 
To overcome these limitations, we propose a novel end-to-end learning scheme to keep track of the
evolution of attrition patterns for the predictive modeling.
It integrates user activity logs, dynamic and static user profiles 
based on multi-path learning. It exploits historical user records by establishing 
a decaying multi-snapshot technique. And finally it employs the precedent user 
intentions via guiding them to the subsequent learning procedure.
As a result, it addresses all disadvantages of conventional methods.
We evaluate our methodology on two 
public data repositories and one private user usage dataset provided by Adobe Creative Cloud.
The extensive experiments demonstrate that it can offer the appealing
performance in comparison with several existing approaches as rated by 
different popular metrics. Furthermore, we introduce an advanced
interpretation and visualization strategy to effectively characterize
the periodicity of user activity logs. It can help to pinpoint important factors that are critical to user attrition and retention and thus suggests actionable improvement targets for business practice.
Our work will provide useful insights into the prediction and elucidation of other user intended actions as well. 
\end{abstract}

\begin{IEEEkeywords}
Customer Attrition, Predictive Modeling, Interpretation
\end{IEEEkeywords}

\section{Introduction}
Being able to predict user intended actions and elucidate underlying behavior patterns 
are of significant value for the business development.
Such intended actions include, but not limited to, user conversion (e.g., purchase, signup), attrition (e.g., churn, dropout), default (failure to pay credit cards or loans), etc. 
These user actions directly lead to revenue gain or loss for companies. 
The capability of predicting user intended actions may help companies to take proactive measures to optimize business outcome. 
In this paper, we focus on predicting attrition, which is one of the most representative user intended actions. 
Attrition, in a broad context, refers to  individuals or items moving out of a collective group over a specific time period\footnote{https://en.wikipedia.org/wiki/Churn$\_$rate, https://www.ngdata.com/what-is-attrition-rate/}.
It can be specialized, as seen in broad applications in different fields. For example, Massive Open Online Courses (MOOCs)\footnote{http://mooc.org} can offer  an affordable and flexible way
to deliver quality educational experiences on a new scale. However, the accompanied high dropout rates are a major concern for educational investors \cite{halawa2014dropout}. 
In the commercial context, the revenue growth of enterprises heavily relies on the acquisition of new customers and retention of existing ones. 
Previous researches and reports have shown that retaining valuable customers is cost effective and more rewarding than acquiring new customers \cite{verbeke2012new, woodruff1997customer}. 
Accordingly, targeting at-risk attrited users in advance and taking intervention measures proactively 
is crucial for improving students' engagement and maintaining customers' retention. 
It helps to sustain the prosperity of MOOCs and enterprises.

There are, however, several inherent challenges confronted 
in predicting attrition using user usage data.
(1) User alignment is a tricky problem as the improper alignment may incur intrinsic bias in the subsequent modeling;
(2) Multi-view heterogeneous data sources, ranging from user activity logs to dynamic and static user profiles,
pose a barrier to the effective interaction and amalgamation;
(3) It is not a trivial task to characterize primitive user activity logs, let alone integrating them with the
downstream predictive modeling effectively and seamlessly;
(4) How to keep track of the evolving intentions of observed historical records for
improving attrition
within a target time period has yet to be explored fully;
(5) It remains unclear how to quantify and visualize the importance of underlying activity patterns, attrition and retention factors.

To address these challenges, we revisit the attrition problem from both predictive modeling 
and underlying patterns representation sides. 
To be specific, we first introduce an appropriate user alignment scheme
based on the calendar timeline, which can
remove the bias as mentioned before. Under an unbiased framework, we propose a 
{\it Blended Learning Approach} (BLA) to address related issues, which renders an appealing predictive performance. 
BLA is mainly characterized as multi-path learning, intention guidance and multi-snapshot mechanism.
The multi-path learning embeds heterogeneous user activity logs, dynamic and static
user information into an unified learning paradigm.
The multi-snapshot mechanism integrates historical user actions explicitly into
the model learning for tracking the evolution of patterns, which is further enhanced by the intention guidance and decay strategies.
For multi-snapshot mechanism, the summarization strategy is developed to 
bridge the separation of the labor-intensive aggregation of user activities and model learning.
The model performance is evaluated on two public data repositories and one dataset of Adobe Creative Cloud user subscriptions.
Furthermore, a simple yet effective visualization approach is introduced to  
discover underlying patterns and to identify attrition and retention factors from user activities and profiles.
This may be exploited by the business or educational units to develop a personalized retention strategy for retaining their users. 



The main contributions and findings of our research are highlighted as follows:

\begin{itemize}

%
%

\item A novel learning scheme is proposed to address several issues involved 
in the user attrition modeling.

\item Comprehensive experiments are performed to evaluate 
the developed methods against baseline approaches and demonstrate the necessity of 
different proposed components. 

\item The periodicity of user historical activities in terms of impacts on future attrition as well
as attrition and retention factors are discovered. 

\end{itemize}

\section{Related Work}\label{Related}

In the past decade, the attrition modeling has been widely studied \cite{vafeiadis2015comparison}. 
Numerous works revolved around on binary classification algorithms. 
The main approach is to build a set of features for users and then train
a classifier for the task. 
Classical data mining algorithms including logistic regression \cite{nie2011credit}, support vector machine (SVM) \cite{coussement2008churn,farquad2014churn} 
and random forest \cite{coussement2008churn,xie2009customer,nagrecha2017mooc}
are intensively studied for attrition prediction. 
Among them, random forest is found to be able to achieve the best performance in many fields like 
the newspaper subscription \cite{coussement2008churn}. 
Actually, random forest is also the modeling algorithm for 
customer behavior analysis including attrition or retention behind predictive analytics 
startup {\it Framed Data}\footnote{https://wefunder.com/framed,http://framed.io} (acquired by Square) \cite{spanoudes2017deep}.
Besides, some biologically inspired methods like genetic programming \cite{idris2012genetic}, 
evolutionary learning algorithm \cite{au2003novel} and vanilla deep neural networks (DNN) \cite{mozer2000predicting,sharma2013neural,tsai2009customer}
were also proposed to search for attrition patterns.  Amongst algorithms of this kind, 
DNN becomes a rapidly growing research direction \cite{spanoudes2017deep,sharma2013neural,tsai2009customer}.
With the growing popularity of deep learning, 
some advanced methods like convolutional neural networks (CNN) \cite{wangperawong2016churn} and 
recurrent neural networks (RNNs) \cite{kasiran2012mobile} 
have been utilized recently as well. 
These works, however, focus on the provided latest attrition status of users.
Essentially, they leave out the evolution of historical states inadvertently.
The precedent statuses would probably be informative in the inference of future statuses by 
coordinating the feature representation better.

There are sporadic works that have been proposed to exploit historical statuses for attrition prediction \cite{wei2002turning,song2006mixed}. 
Although these works have tried to incorporate historical statuses of users, they have two issues.
First, the whole historical observation periods 
are divided into multiple sub-periods for model training with 
handcrafted efforts. In this case, the correlation
across different sub-periods cannot be fully explored. 
Second, the decaying impact of statuses within different sub-periods
on attrition prediction within the target time period is out of consideration. 
The survival analysis framework has been proposed to capture the time-to-event of attrition \cite{lu2002predicting}.
It utilizes the initial information at the start of the user enrollment to perform model learning 
for predicting survival time of subscriptions. 
Here the inherent problem is that the evolving user activities 
are not incorporated into the attrition prediction, which
are crucial to the attrition modeling according to our experiments.
Aside from the above works purely based on attrition, 
profit-driven discussion and simulation studies were also 
performed based on a potential intervention assumption (e.g., bonus, discount) \cite{mozer2000predicting}.

Compared with the intensive research on predictive modeling, 
little work focuses on the interpretation of attrition prediction results in terms of at both individual and class/group level. 
This is in part due to inherent challenges faced by non-interpretable classifiers under the framework 
of traditional interpretation methods \cite{ribeiro2016model,nagrecha2017mooc}. 
Recently, advanced interpretation methods like saliency maps \cite{simonyan2013deep} and its follow-up work Local Interpretable Model-Agnostic Explanations (LIME) \cite{ribeiro2016should} have
been proposed in this regard. Our technical approach to distilling attrition insights is inspired by saliency maps.

\section{Modeling}\label{Modeling}

\subsection{Preliminary}\label{Preliminary}

\begin{figure}[!h]
\centering
\includegraphics[height=0.6\columnwidth, width=\columnwidth]{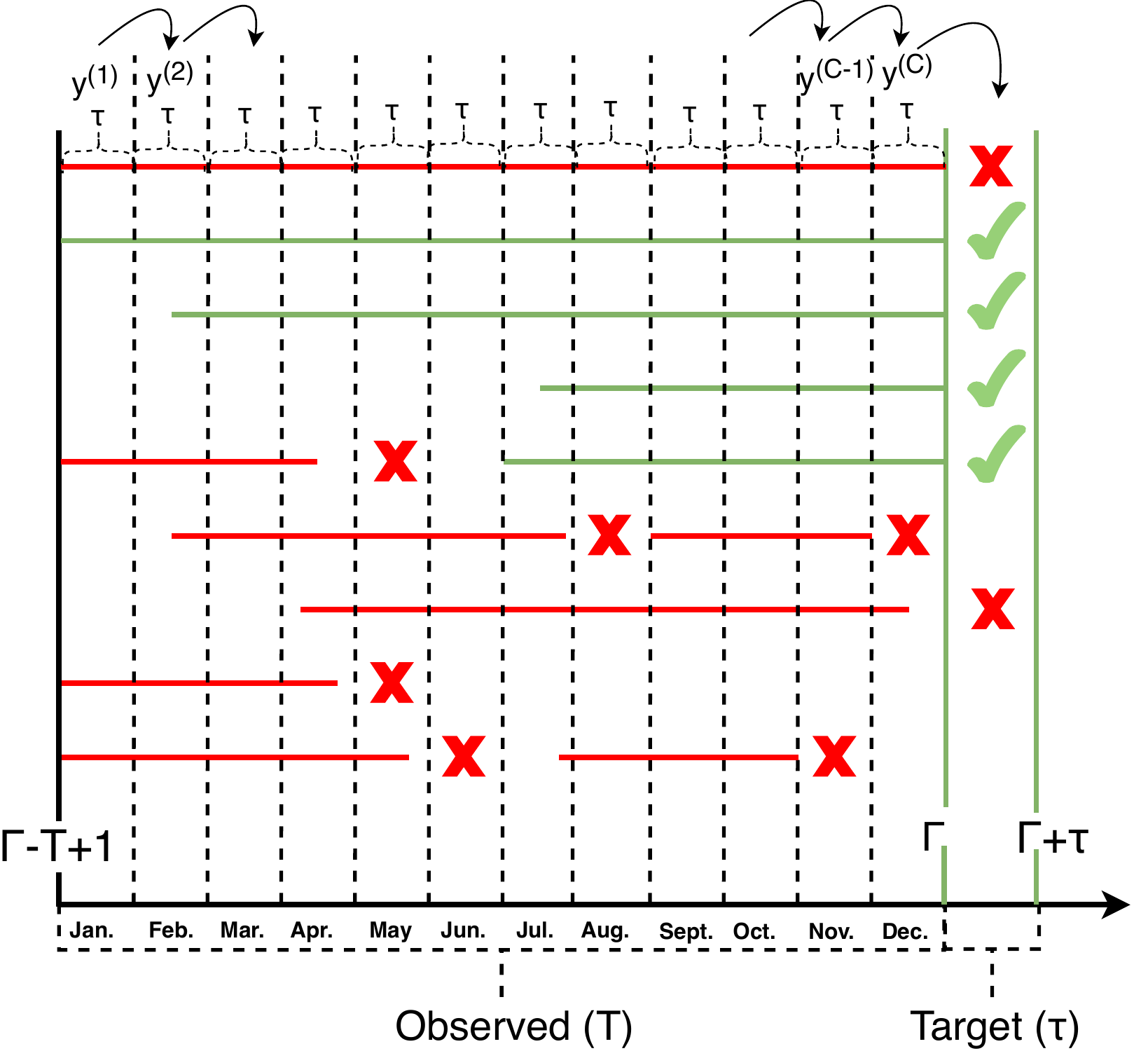}
\caption{Schematic overview of different user statuses with varied types of observed activity logs.
There are $C=T/\tau$ snapshots.
$\times$ and $\checkmark$ denote attrition and retention, respectively. 
$\curvearrowright$ indicates that the ground-truth label of user activities during snapshot $t-1$ is the attrition status within future snapshot $t$, $2 \leq t \leq C$.}
\label{Terminate}
\end{figure}


In this section, we focus on formulating the attrition prediction problem.
To facilitate the problem formulation, we give a schematic illustration of user statuses as shown in Fig. \ref{Terminate}. 
Suppose there are a set of N samples or users $\mathbb{D} = \{(\mathbf{X}_i, y_i)\}_{i=1}^N$, for which we collect user data $\mathbf{X}_i$ 
from the historical time period [$\Gamma$-T+1, $\Gamma$] of length T, and we aim to predict their statuses $y_i \in \mathcal{Y} = \{0, 1\}$ 
in the future target time window [$\Gamma$+1, $\Gamma$+$\tau$] of length $\tau$\footnote{Theoretically speaking, $\tau$ is flexible and can be any positive integer.
In practice, it depends on business scenarios, which can be weekly, biweekly, monthly or longer periods.}.

The user data $\mathbf{X}_i$ are composed of three primitive heterogeneous sub-components: 
activity logs on a basis of observed time granularity (e.g., day) $\mathbf{X}_{ia}$, dynamic user information $\mathbf{X}_{id}$, and static user profiles $\mathbf{X}_{is}$, 
namely, $\mathbf{X}_i = (\mathbf{X}_{ia}, \mathbf{X}_{id}, \mathbf{X}_{is}) \in \mathcal{X}$. For the activity logs component, 
we denote any events happening during the time span T right prior to the target time window as 
$\mathbf{X}_{ia} = \Big(\mathbf{X}_{ia}^{(\Gamma - T + 1)}, \mathbf{X}_{ia}^{(\Gamma - T + 2)}, \ldots, \mathbf{X}_{ia}^{(\Gamma - 1)}, \mathbf{X}_{ia}^{(\Gamma)}\Big)$.

The general goal is to search for a reasonable mapping rule from observed feature space to attrition statuses $\mathcal{R}(\cdot): \mathcal{X} \rightarrow \mathcal{Y}$ and subsequently apply $\mathcal{R}(\cdot)$ to 
estimate the statuses of samples in the future. The probability of sample $i$ in attrition can be denoted as
\begin{equation}
p(y_i =1 | X)
\end{equation}
Practically speaking, the ground truth is relatively subject to the future target time window [$\Gamma+1$, $\Gamma+\tau$]. 
Specifically, if a user drops out of a course or is churned within this window, it is then labeled as $1$; 
if the user remains active, then it is labeled as 0.
It is worth noting that attrition labels are generated based on the overall statuses of users during the target time window.

\begin{figure}[!h]
\centering
\includegraphics[height=0.635\columnwidth, width=1\columnwidth]{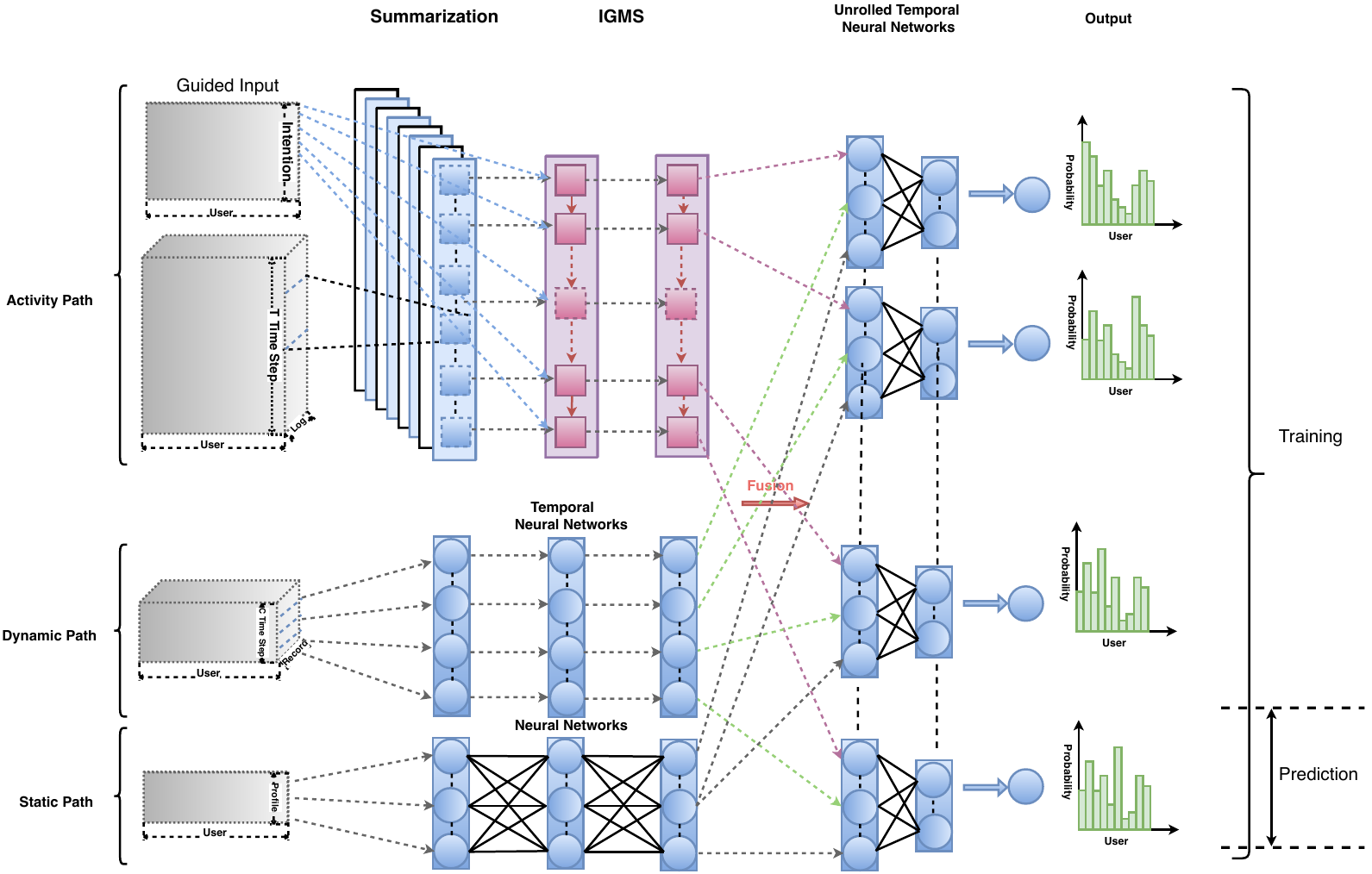}
\caption{Framework overview of Blended Learning Approach}
\label{Framework}
\end{figure}

\subsection{Methodology}\label{Methodology}
Section \ref{Preliminary} introduces the primitive problem formulation. 
We propose to extend this formulation to incorporate multi-snapshot statuses according to the
snapshot window, which is equal to the pre-designated target time window size $\tau$. 
Concretely, sequential outputs are generated across sampled observed time period per $\tau$ units 
based on the attrition definition.
We then can generate $C = \frac{T}{\tau}$ snapshot outputs. 
As for users with the observed time span being less than $T$, 
we take zero-padding for the computational convenience. 
The corresponding masking indicators are introduced to disable their contributions to the loss as detailed in Eq. \ref{J}.
Accordingly, we obtain the final series of statuses of sample $i$ as $\Big(y_{i}^{(1)}, y_{i}^{(2)}, \ldots, y_{i}^{(C)}\Big)$ where $y_{i}^{(C)}$ is the status within the target time period. 
In this case, the conditional probability that sample $i$ is in the state of attrition can be represented as

\begin{equation}
p\Big(y_i^{(t)} =1 | X; y_i^{(t-1)}, \ldots, y_i^{(1)}\Big), 2 \leq t \leq C
\end{equation}
Therefore, our learning rule can naturally evolve to be 
$\mathcal{R}(\cdot): (\mathcal{X}, \mathcal{Y}^{t-1}) \rightarrow \mathcal{Y}$ for target time step $t$.

With the reformulation of this problem, 
we introduce different learning layers/components of 
BLA and discuss how these components tackle the 
aforementioned issues faced by the attrition prediction.

\subsubsection{\textbf{Parallel Input Layer}}

In accordance with reformulated mapping rule $\mathcal{R}(\cdot)$, 
the original feature space includes four different parts:
activity logs, dynamic information, static profiles
and precedent statuses. 
In this case,  
we design multiple parallel input layers for corresponding learning paths to 
solve the amalgamation problem associated with 
these heterogeneous multi-view features as diagrammed in Fig. \ref{Framework}.

{\it \textbf{Activity input layer}}--
Three-dimensional users activity logs are fed into this layer, 
along which are user samples, observation time span, and 
activity metrics. 
Concretely, the granularity of primitive observation time 
can be, but not limited to, every minute, hourly, daily, weekly, monthly, or any reasonable time duration.
The activities can be, but not limited to, students' engagement for MOOCs, 
products booting, usage of specific features within the products for software companies.
{\it \textbf{Dynamic input layer}}--
Three-dimensional dynamic layer is responsible for the derivative of the user profile, products information or their interactive records 
based on the snapshot window. This includes, but not limited to, subscription age, payment settings (automatic renewal/cancellation), or
any reasonable derivatives.
{\it \textbf{Static input layer}}--
This layer takes static profiles of users or products, which cover many details including
but not limited to, gender, birthday, geographical location, market segments, registration/enrollment method or
any other unchanging information. 
This layer is of the two-dimensional shape.
{\it \textbf{Guided input layer}}--
The snapshotted statuses as a two-dimensional guided intention is embodied into the attrition prediction 
through this layer.

\subsubsection{\textbf{Summarization Layer}}

Closely following the activity input layer is the summarization layer,
which is developed for summarizing user activities.
Due to the homogeneity along the observed time and the heterogeneity across activity logs, 
we utilize one-dimensional CNN to aggregate low-level activity logs over a fine-grained time span (e.g., day)
to generate high-level feature representation over a coarse-grained one (e.g., week).
Mathematically speaking, we have

\begin{equation}
f_s^{(t)}(\mathbf{X}_{ib})= \sum_{m=0}^{M-1}\sum_{n=0}^{N-1}\mathbf{W}_{mn}^s\mathbf{X}_{ib}^{(t+m),(n)}
\end{equation}
where $\mathbf{X}$ is the input activity logs, 
$t$ and $s$ are the indices of output time step and 
activity summarizer, respectively. 
Summarizer $\mathbf{W}^s$ is the $M \times N$ weight matrix with $M$
and $N$ being the window size of summarizing time span and sequence channel, respectively. 
In particular, N of the first summarization layer is equal to the number of activity metrics. 
Activity logs can be summarized to be with different granularities via setting kernel size $M$. 

The designed summarization layer entails threefold benefits:
(1) Learning rich relations and bypassing labor-intensive handcrafted efforts in summarizing primitive activity logs;
(2) Upholding the interpretation track of primitive activity metrics compared with hand-operated aggregation;
(3) Accelerating the training procedure of model thanks to the noise filtering and feature dimensionality reduction.

\subsubsection{\textbf{Intention Guided LSTM Layer with Multiple Snapshot Outputs}}
In order to capture the long-range interactive dependency of 
summarized activities and make the most use of generated auxiliary statuses, 
we propose to introduce a variant of Long Short-Term Memory Networks (LSTM) \cite{hochreiter1997long}.
To simplify the following notations, we omit sample indices here.
The original formulation in the family of Recurrent Neural Networks \cite{rumelhart1985learning,Goodfellow-et-al-2016} (RNNs)\footnote{http://colah.github.io/posts/2015-08-Understanding-LSTMs/} 
is usually denoted as 
\begin{equation}
h^{(t)} = f\big( h^{(t-1)}, x^{(t)}\big)
\end{equation}
where $x^{(t)}$ and $h^{(t)}$ are the input sequence of interest and the estimated hidden state vector or output at time $t$, respectively. 
$h^{(t-1)}$ is the immediate precedent estimated state vector.
We here propose to embed the actual immediate precedent status $y^{(t-1)}$ 
to guide the learning procedure as
\begin{equation}
h^{(t)} = f\big( h^{(t-1)}, x^{(t)}, y^{(t-1)}\big)
\end{equation}
As illustrated in Fig. \ref{ILSTM}, the core equations are accordingly updated as follows:
\begin{equation}
\begin{split}
f_t &=  \sigma \Big(W_f  \big[h^{(t-1)}, x^{(t)}, y^{(t-1)}\big] + b_f \Big) \\
i_t & = \sigma \Big(W_i  \big[h^{(t-1)}, x^{(t)}, y^{(t-1)}\big] + b_i \Big) \\
o_t & = \sigma \Big(W_o  \big[h^{(t-1)}, x^{(t)}, y^{(t-1)}\big] + b_o \Big) \\
C_t & =  f_t \circ C_{t-1} + i_t \circ \mathrm{tanh} \Big(W_C  \big[h^{(t-1)}, x^{(t)}, y^{(t-1)}\big] + b_C \Big) \\
h_t & = o_t \circ \mathrm{tanh}(C_t) 
\end{split}
\end{equation}
where $\circ$ denotes the element-wise Hadamard product.
$\sigma$ and $\mathrm{tanh}$ are sigmoid and hyperbolic tangent activation functions, respectively. 
$f_t$, $i_t$, $o_t$, $C_t$ are forget, input, output and cell states, which control the 
update dynamics of the cell and hidden outputs jointly.
It is noted that multiple snapshot outputs in the training phase can
keep track of the evolution of statuses sequentially and naturally.
In the meantime, the introduced auxiliary statuses 
are complementary to activity, dynamic and static inputs in terms of 
capturing the intention progression. We call it {\it IGMS} as annotated in Fig. \ref{Framework}.


\subsubsection{\textbf{Temporal Neural Network Layer}}
In order to guarantee the temporal order preservation of feature representation,
we introduce temporal neural networks:
\begin{equation}
a_l^{(t)} = \sigma ( W_{l}^{(t)} a_{l-1}^{(t)} + b_{l}^{(t)} )
\end{equation}
where $a_{l-1}^{(t)}$ is a temporal slice of the output of layer $l-1$.

This layer entails twofold roles: 
(1) Feature learning over different snapshot periods in the dynamic path;
(2) Fusion of feature representation in multiple paths.
It is also noted that the activation function of the final temporal neural network layer is {\it sigmoid}.

\subsubsection{\textbf{Decay Mechanism}}
\begin{figure}[!t]
\centering
\includegraphics[height=0.45\columnwidth, width=\columnwidth]{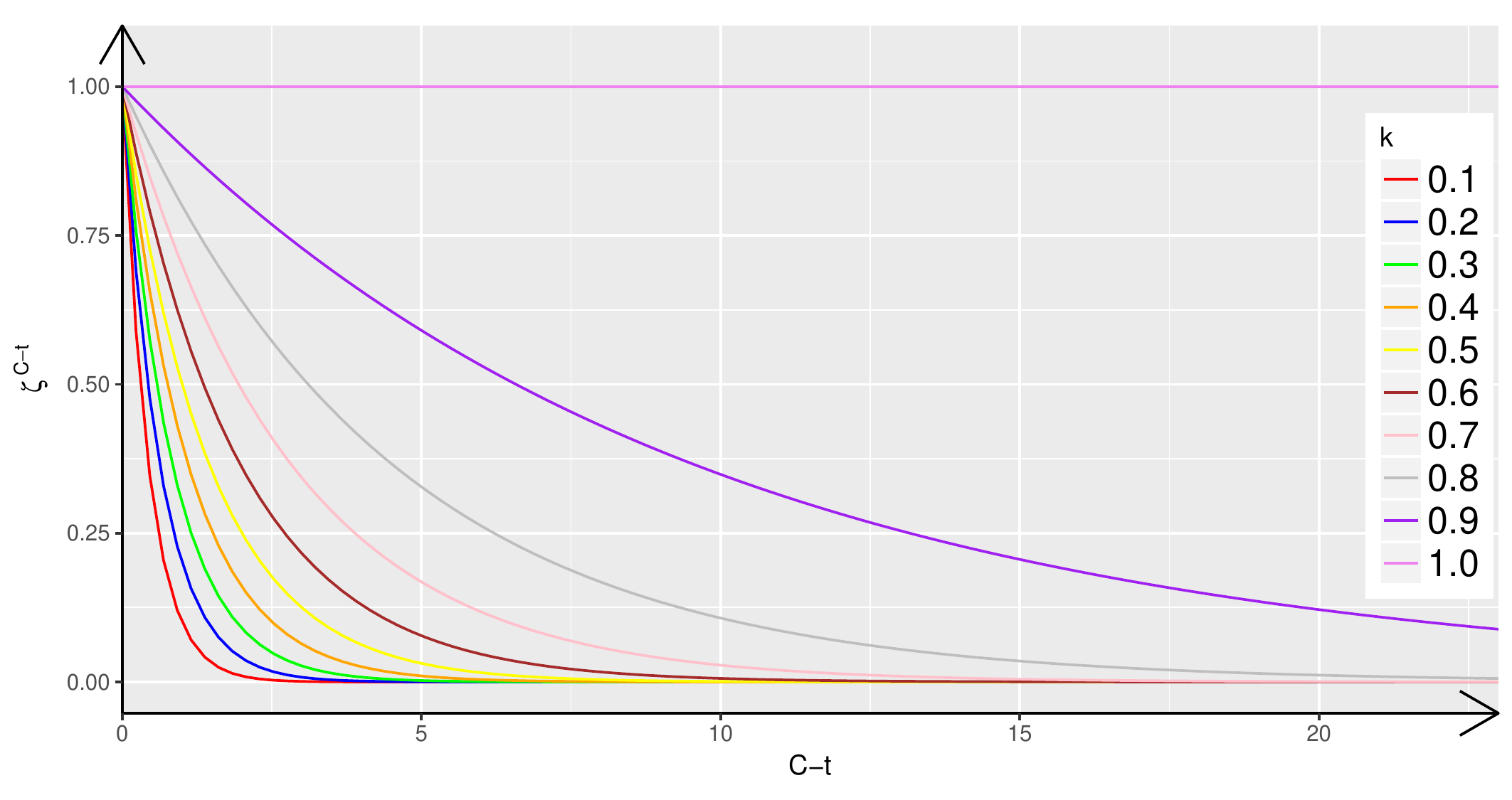}
\caption{Examples of temporal decay of sample weights across different snapshot time steps.}
\label{decay}
\end{figure}

When multiple snapshot attrition statuses are 
incorporated into our learning framework, 
their associated impacts need to be adjusted in the training phase accordingly.
This results from the fact that the underlying behavior patterns might change over time in a certain way \cite{tan2017time}.
To this end, we have a underlying assumption: the bigger the time gap between auxiliary snapshot statuses and 
attrition status at target time period is, the less similar the underlying intention patterns are.
The temporal exponential decay is thus introduced to penalize weights based on this assumption. 
Concretely, $\zeta^{(C - t)} = k^{(C-t)}$, where $k \leq 1$ depends on the expected speed of decay,
as shown in Fig. \ref{decay}. 
Since the decay speed $k$ is a hyper-parameter, it is determined by the validation dataset.

\subsubsection{\textbf{Objective Function}}

\begin{figure}[!t]
\centering
\includegraphics[height=0.4\columnwidth, width=\columnwidth]{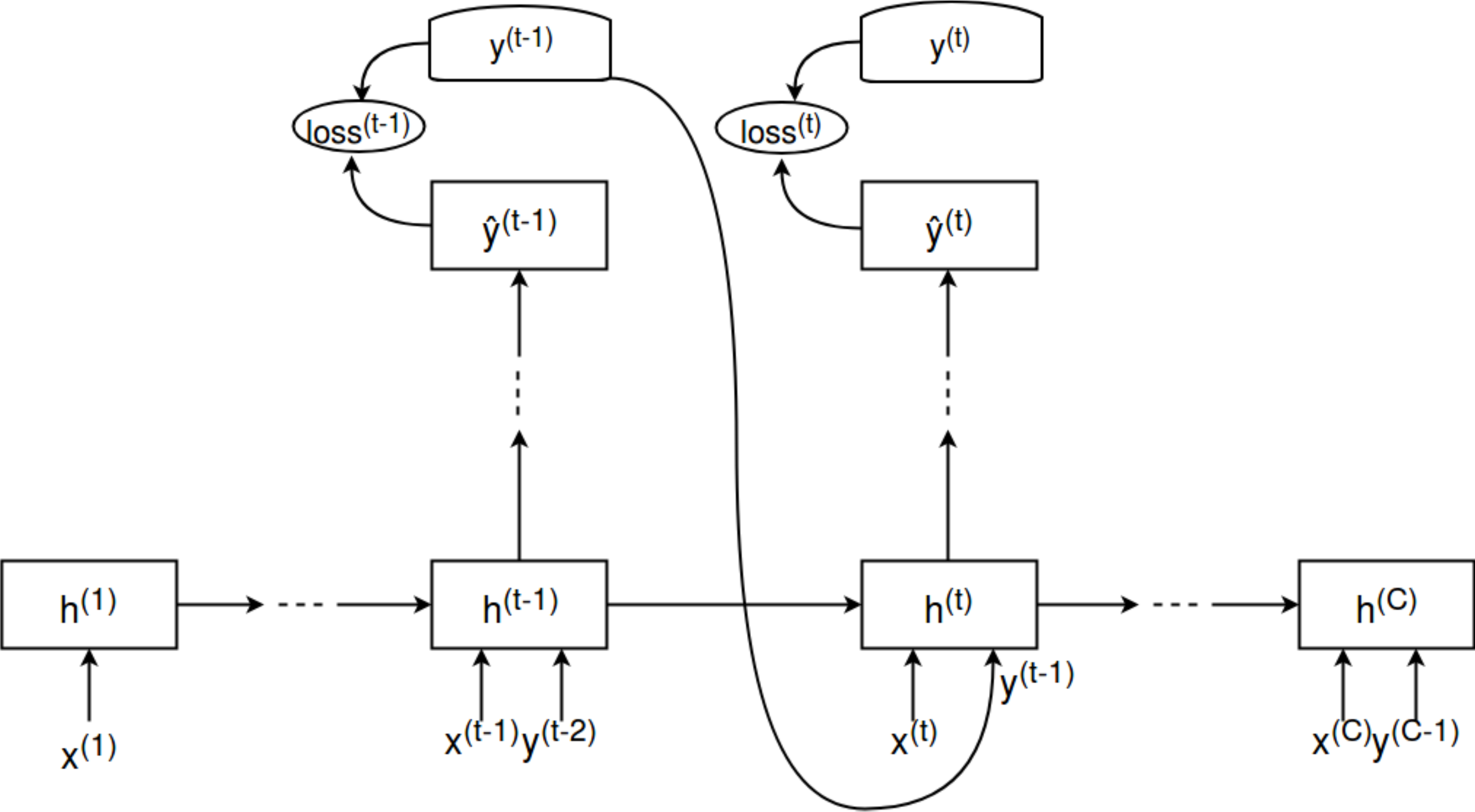}
\caption{Illustration of the guided intention mechanism,
where the precedent actual intentions are also used for attrition estimation in next time step. }
\label{ILSTM}
\end{figure}

With auxiliary snapshot statuses being incorporated into the training phase as shown in Figs. \ref{Framework} and \ref{ILSTM}, 
we have the following loss function to guide the learning procedure:
\begin{equation}
\begin{split}
\mathcal{J} = -\frac{1}{N} \sum_{t=1}^C \zeta^{(C - t)} \sum_{i=1}^N \eta_i^{(t)} \Big[ y_i^{(t)}\mathrm{log}\ p(\hat{y}_i^{(t)} = 1) \\
\ + \ (1-y_i^{(t)})\mathrm{log} \ p(\hat{y}_i^{(t)}=0) \Big]
\label{J}
\end{split}
\end{equation}
where $\zeta^{(t)}$ and $\eta_i^{(t)}$ are temporal decay weight and sample-level binary masking indicator, respectively.
In particular, $\eta_i^{(t)}$ can be used to mask invalid attrition statuses of training samples in the snapshot time periods caused by
the calendar date alignment. For example, the registration dates of some users are later than the beginning of observed time periods 
as shown in Fig. \ref{Terminate}.

As shown in Fig. \ref{Framework}. 
BLA mainly includes activity path, dynamic path and static path. 
In the activity path, the time granularity of input is on a basis of 
primitive observed time granularity (e.g., day), whereas output granularity is the snapshot window (e.g., month).
The dynamic path is composed of temporal neural networks with both input and output being at the granularity of the snapshot span.
For the static path, the outputs are forked $C$ times for further fusion with outputs of activity and static paths, 
as shown in unrolled temporal neural networks of Fig. \ref{Framework}.

\subsubsection{\textbf{Predictive Inference}}\label{Inference}
With the learning architecture and estimated parameters, 
we obtain the learned model ready for predicting user-intended actions. 
As illustrated in Fig. \ref{Framework}, 
we have only one output at the $C^{th}$ time period, which is the attrition probability 
of the target time period in prediction phase (validation and test).

\subsection{Feature Interpretation and Visualization}\label{viz}
Saliency maps are one powerful technique to interpret and visualize feature representation behind deep neural networks, which
have been widely utilized to analyze feature importance \cite{simonyan2013deep, tan2018deep}.
In this paper, we also construct saliency maps by back-propagating features with the guidance of BLA  to highlight 
how input features impact the user attrition. 
First of all, a user is supposed to have feature vector $x_0$ and the associated attrition state, 
we aim to figure out how elements of $x_0$ shape output probability of state $\mathcal{R}(x_0)$.  Regarding BLA, the score $\mathcal{R}(x)$ is 
a highly non-linear function of input $x$. $\mathcal{R}(x)$, however, 
can be approximated by a linear function in the closeness of $x_0$ based on the first-order Taylor expansion:
\begin{equation}
\mathcal{R} (x) \approx w^T(x - x_0) + \mathcal{R} (x_0)
\end{equation}
where $w$ is the first-order derivative of $\mathcal{R} (x)$ with respect to the feature vector $x$ at $x_0$:
\begin{equation}
w = \frac{\partial{\mathcal{R}(x)}}{x}\Big|_{x_0}
\end{equation}
There are two points about the interpretation of this kind to consider: 
1) The magnitude of the derivative indicates the extent to which the change of the most influential elements 
of feature vector on the probability of the attrition state;
2) The direction of each element of the derivative shows whether such a change boosts or decreases the probability of the attrition state.
It is noted that the computation of the user-specific saliency map is very fast due to the requirement of a single back-propagation pass.

For dynamic and static inputs, we take average on saliency maps of all test users and then obtain the overall saliency map. 
The overall one can help to identify the underlying attrition and retention factors involved in the attrition directly. 
For activity logs with different metrics, 
we concentrate on exploring the evolution patterns of activity logs. 
Thus, we take the absolute value of saliency maps before averaging over all test users.
Finally, we take sum of all metrics along observed time periods.


\begin{table*}[!t]
\centering
\caption{Basic statistics of MOOCs and KKBox. Observation span T and snapshot window size $\tau$ are detailed in section \ref{Preliminary}.
$\tau$ is set based on business scenarios (e.g., subscription plan) here.}
\resizebox{\textwidth}{!}{%
\begin{tabular}{ c|c|c|c|c|c|c}
\hline 
& \# of user & \#  of attrition & \# of persistence & observation span T (day) & snapshot window size $\tau$ (day) &  target time period  \\ \hline
\multirow{1}{*}{MOOCs}  
& 120,542	& 95,581      & 24,961  &   30  &  10 &10 days after the end of observed days  \\ 
\hline
\multirow{2}{*}{KKBox}  
& 11,2118	&   19,415     & 92,703   &  720 & 30 & 02/01/2017 $\sim$ 02/28/2017    \\ 
&  156,029	&    21,752     &    134,277 &   720 & 30 & 03/01/2017 $\sim$ 03/31/2017    \\ 
\hline 
\end{tabular}
}
\label{Mooc_KKBox_Statistics}
\end{table*}

\section{Experiment}\label{Experiment}
In this section, we first assess the performance of BLA on the customer attrition task comparing with 
competitive baselines for two public datasets and one private dataset. 
Then, we perform feature analysis to distill the evolving patterns of user activity logs, attrition and retention factors.

\subsection{Experimental Setup}
We utilize python libraries Keras\footnote{https://github.com/keras-team/keras} to build the architecture of our learning algorithm
and Tensorflow \cite{abadi2016tensorflow} to perform feature interpretation and visualization. 
NVIDIA Tesla K80 GPU with memory of 12GB is used for model development. 
Microsoft Azure with PySpark is adopted as the large-scale data processing platform. 

\textbf{Network Architecture}. Along 
{\it Activity Path} are 1 one-dimensional CNN (14 kernels)
and 2 intention-guided LSTM (30 and 15 kernels).
{\it Dynamic Path} consists of 1 two-layered temporal neural networks with 30 and 15 hidden nodes.
{\it Static Path} involves 1 two-layered neural networks with 30 and 15 hidden nodes.
The fusion layer includes 1 two-layered temporal neural networks with 30 and 15 hidden nodes.

\textbf{Training}.
We initialize parameters in BLA with 
{\it Glorot} uniform distribution 
\cite{glorot2010understanding}.
The mini-batch size and 
the maximum number of epochs
are set to be $128$ and $500$, respectively. 
The parameters are updated based on Adam 
optimization algorithm \cite{kingma2014adam} with 
learning rate of $0.001$ and decay factor of 1e-3.
Early stopping of 20 epochs is set to prevent the overfitting. 
Trainable parameters and hyper-parameters are tuned based on the loss of
attrition records in the validation dataset. 
As shown in Fig. \ref{Framework}, all historical records 
are incorporated into the loss function $\mathcal{J}$ in formula (\ref{J}). 

\textbf{Test}. 
As shown in Fig. \ref{Framework},
the prediction is conducted on customer attrition records during the target time periods. 
With both the trained parameters and hyper-parameters, we measure the performance 
of the model on the specified target periods. 

Training and test parts are split based on the temporal logic and 
will be detailed in the coming subsections accordingly.

\subsection{Baseline Approaches}\label{Baselines}
In this section, we will introduce alternative algorithms as baseline schemes to 
demonstrate the effectiveness of the proposed BLA. User activity logs are manually 
aggregated and then reshaped to be a vector. The one-month 
dynamic and static information are directly reshaped and then fused with logs vector to
generate the learning features. The baselines are tuned based on the validation part and the optimal parameters are reported accordingly. 
\begin{itemize}
\item \textbf{LR}: The classical Logistic Regression \cite{nie2011credit} is commonly used with good interpretation capacity \cite{nagrecha2017mooc}.
To facilitate the training with the large-scale dataset, we construct a simple neural network with one input layer and {\it sigmoid} activation function with GPU acceleration.
Adam \cite{kingma2014adam} with learning rate of 0.01 and decay rate of $10^{-3}$ is adopted as the optimization algorithm for MOOCs and KKBox.

\item \textbf{DNN}: Generally speaking, stacking computational units can represent 
any probability distributions in a certain configured way \cite{bengio2009learning}. Thus, 
vanilla deep neural networks are widely utilized for attrition prediction in the academic research \cite{sharma2013neural,tsai2009customer}.
The network is with 2 hidden layers of 100, 10 nodes, respectively. 
Adam \cite{kingma2014adam} with learning rate of 0.01 for MOOCs and 0.001 for KKBox, as well as decay rate of $10^{-3}$ for both datasets is adopted.

\item \textbf{RF}: Random Forest is frequently used in churn or dropout prediction \cite{coussement2008churn,xie2009customer,nagrecha2017mooc} 
and deployed in the industry (e.g., Framed Data), which usually shows a good performance \cite{spanoudes2017deep}.
The RandomForest of Scikit-Learn with the tree number of 20 and the maximum depth of 30 is employed for MOOCs and KKBox.

\item \textbf{NB}: Naive Bayes \cite{nath2003customer,huang2012customer,zhang2004optimality}
is also explored in the user attrition prediction. 
We adopt the classical Gaussian Naive Bayes algorithm for the classification. 

\item \textbf{SVM}: SVM is explored in this regard as well \cite{coussement2008churn,farquad2014churn}. 
To scale better to large numbers of samples (the inherent problem in SVM training),
we adopt liblinear (LinearSVC) for {\it linear} kernel,  the bagging classifier (BaggingClassifier + SVC) for non-linear {\it radial basis function (rbf)} and {\it polynomial (poly)} kernels
in Scikit-Learn library. The settings are linear kernel with $C=0.001$ for MOOCs and rbf kernel with $C=0.001$ for KKBox.


\item \textbf{CNN}: Convolutional neural networks \cite{wangperawong2016churn} include
two layers of one-dimensional convolutional neural networks with 14 and 7 kernels 
and the subsequent fully connected neural networks 30 and 15 hidden nodes.

\item \textbf{LSTM}:
Vanilla recurrent neural networks or long short-term memory networks \cite{kasiran2012mobile,fei2015temporal}
are also utilized here by aggregating activity logs with handcrafted efforts.
One two-layered LSTM with 30-dimensional and 15-dimensional output nodes, followed by
subsequent fully connected neural networks with 30 and 15 hidden nodes.
\end{itemize}

The variants of BLA are listed as follows:
{\it MSMP}: a variant without intention guidance;
{\it IGMP}:  a variant without multi-snapshot mechanism;
{\it IGMS-AD}: a variant only using activity path and dynamic path;
{\it IGMS-AS}: a variant only using activity path and static path;
{\it IGMS-DS}: a variant only using dynamic path and static path.

\subsection{Evaluation Metrics}
To measure the prediction performance of the proposed methodology,
we adopt F1 Score, Matthews correlation coefficient (MCC) \cite{matthews1975comparison},
the Area under Curves of Receiver Operating Characteristic (AUC@ROC) \cite{tan2014link,tan2016modeling,tan2018modeling} and Curves of
Precision-Recall (AUC@PR), respectively. As opposed to ROC curves, Precision-Recall curves are more 
sensitive to capture the subtle and informative evolution of algorithm's performance. 
A more in-depth discussion is detailed in Ref. \cite{davis2006relationship}.

\subsection{Experimental Results}

We perform attrition prediction on two public attrition repositories
MOOCs (dropout prediction)\footnote{https://biendata.com/competition/kddcup2015/}  
and KKBox (churn prediction)\footnote{https://www.kaggle.com/c/kkbox-churn-prediction-challenge} based on
BLA against baseline approaches. 
Furthermore, we apply the proposed method to users of Adobe Creative Cloud (CC) and 
compare it with random forest and the currently deployed model.  

\subsubsection{MOOCs and KKBox}

Dropout in MOOCs and churn in subscription-based commercial products or services are two typical scenarios 
associated with the attrition problem. 
Dropout prediction focuses on the problem where we prioritize  
students who are likely to persist or drop out of a course, which 
is usually characterized by the highly skewed dominance of dropout over persistence. 
As opposed to dropout in MOOCs, churned users are in tiny proportion compared with persistent
ones.  The basic statistics of the MOOCs and KKBox datasets are described in Table \ref{Mooc_KKBox_Statistics} briefly.
Here attrition labels indicate the user status within the target time period. 
As the given spans of the target time period are 10 days for MOOCs and one month for KKBox\footnote{$\tau$ is pre-specified in datasets here.}, we set snapshot span as $\tau = 10$ and $\tau = 30$ accordingly. 
Given observation span $T = 30$ and $T = 720$, a total of $C=3$ and $C=24$ outputs are 
generated simultaneously. The last one is the status to predict, and the precedent outputs  
are auxiliary statuses for aiding in the model development. 
User activity logs, dynamic and static features are given 
in Tables \ref{Mooc_KKBox_Profile} and  \ref{Mooc_KKBox_Information}, respectively.
For MOOCs, the stratified data splitting is adopted since 
there are few overlapping time spans among different courses.
Accordingly, the ratio of training, validation and testing datasets is 6:2:2.
For KKBox, user records on Feb 2017 and March 2017 are utilized as 
model development and assessment, respectively. The development dataset
is further split into internally stratified training and validation parts with ratio 8:2. 

The comparison results among BLA and baselines are examined in Tables \ref{Moocs_Metrics} and \ref{KKBox_Metrics}.
Overall, BLA is able to outperform other commonly used methods for attrition prediction in terms of an array of metrics. 
In MOOCs, we report F1 score and AUC@PR based on minor persistent users,
which is sensitive to the improvement of algorithms. 
It is noted that, as compared to baselines, the performance gain of BLA is more obvious in KKBox than that in MOOCs. 
There are two underlying causes: 
(1) Few dynamic and static user features are available for MOOCs, 
which degrades the power of the multi-path learning as shown in Table \ref{Mooc_KKBox_Information};
(2) The span of historical records is limited for MOOCs, which will suppress 
the multi-snapshot mechanism inevitably as shown in Table \ref{Mooc_KKBox_Statistics}.




\begin{table}[!h]
\centering
\caption{Student engagement logs of MOOCs and customer music listening logs of KKBox.}
\resizebox{0.5\textwidth}{!}{%
\begin{tabular}{ c|c|c }
\hline
 Dataset & Activity  & Remarks \\ \hline
\multirow{6}{*}{MOOCs}  
& problem  &  working on course assignments \\ 
& video  &  watching course videos \\ 
& access   &  accessing other course objects except videos and assignments \\ 
& wiki  &  accessing the course wiki \\ 
& discussion  &   accessing the course forum \\ 
& navigate  &   navigating to another part of the course \\ 
& page\_close  &  closing the web page \\ 
& source  &   Event source (server or browser) \\ 
& category  &   the category of the course module \\ 
\hline
\multirow{7}{*}{KKBox}  
&  num\_25  &  \# of songs played less than 25\% of the song length  \\ 
&  num\_50  & \# of songs played between 25\% to 50\% of the song length   \\ 
&   num\_75 &  \# of songs played between 50\% to 75\% of the song length  \\
&  num\_985  &  \# of songs played between 75\% to 98.5\% of the song length  \\ 
&  num\_100  &  \# of songs played over 98.5\% of the song length   \\ 
&   num\_unq &  \# of unique songs played   \\ 
&  total\_secs   &  total seconds played \\ 
\hline
\end{tabular}
}
\label{Mooc_KKBox_Profile}
\end{table}

\begin{table}[!h]
\centering
\caption{User dynamic and static profile of MOOCs and KKBox.}
\resizebox{0.5\textwidth}{!}{%
\begin{tabular}{ c|c|c|c }
\hline
Dataset & Style  & Profile & Remarks \\ \hline
\multirow{2}{*}{MOOCs}  
& \multirow{1}{*}{dynamic}  &  ---   &  ---  \\ 
& \multirow{1}{*}{static}  & course\_id  &  course ID   \\ 
\hline
\multirow{7}{*}{KKBox}
& \multirow{3}{*}{dynamic}  
&  membership  &   the time to the initial registration   \\ 
&&   is\_auto\_renew &  whether subscription plan is renewed automatically  \\ 
&&   is\_cancel  &   whether subscription plan is canceled \\ 
\cline{2-4}
& \multirow{4}{*}{static}  
&  bd  &  age when registered  \\ 
&&  city  &  city  when registration  (21 anonymous categories) \\ 
&&  gender &  gender (male and female)  \\
&&  registered\_via  &   registration method (5 anonymous categories) \\ 
\hline
\end{tabular}
}
\label{Mooc_KKBox_Information}
\end{table}

\begin{table}[!h]
        \centering
        \caption{Performance comparison on MOOCs for attrition prediction.}
        \label{Moocs_Metrics}
        \resizebox{\columnwidth}{!}{%
        \begin{tabular}{ccccc}
        \hline
        \multicolumn{1}{c|}{\textbf{Method}} & \textbf{AUC@ROC} & \textbf{MCC} & \textbf{F1 Score} & \textbf{AUC@PR}  \\ \hline

\multicolumn{1}{c|}{LR}	&	0.8595 	&	0.5477  	&	0.6017  	&	0.7003	\\ 

\multicolumn{1}{c|}{RF}	&	0.8693  &	0.5753  	&	 0.6430 	&	 0.7077 	 \\ 

\multicolumn{1}{c|}{DNN}	&	 0.8718 	&	0.5786  	&	0.6509 	&	 0.7157  \\ 

\multicolumn{1}{c|}{NB}	&	0.8354  	&	 0.4976  	&	0.5925	&	0.6562   \\ 

\multicolumn{1}{c|}{SVM}	&	 0.8656	&	 0.5440  	&	 0.5924 	&	0.7049  \\ 

\multicolumn{1}{c|}{CNN}	&	 0.8778  	&	  0.5851  	&	0.6480   	&	0.7324     \\ 

\multicolumn{1}{c|}{LSTM}	&	0.8746  &	  0.5863	&	 0.6528	&	0.7246   \\ \hdashline


\multicolumn{1}{c|}{BLA}	&	\textbf{0.8842}	&	\textbf{0.5973}	&	\textbf{0.6569}	&	\textbf{0.7464}	 \\ \hline 

\end{tabular}%
        }
\end{table}

\begin{figure}[!h]
\centering
\includegraphics[height=0.5\columnwidth, width=\columnwidth]{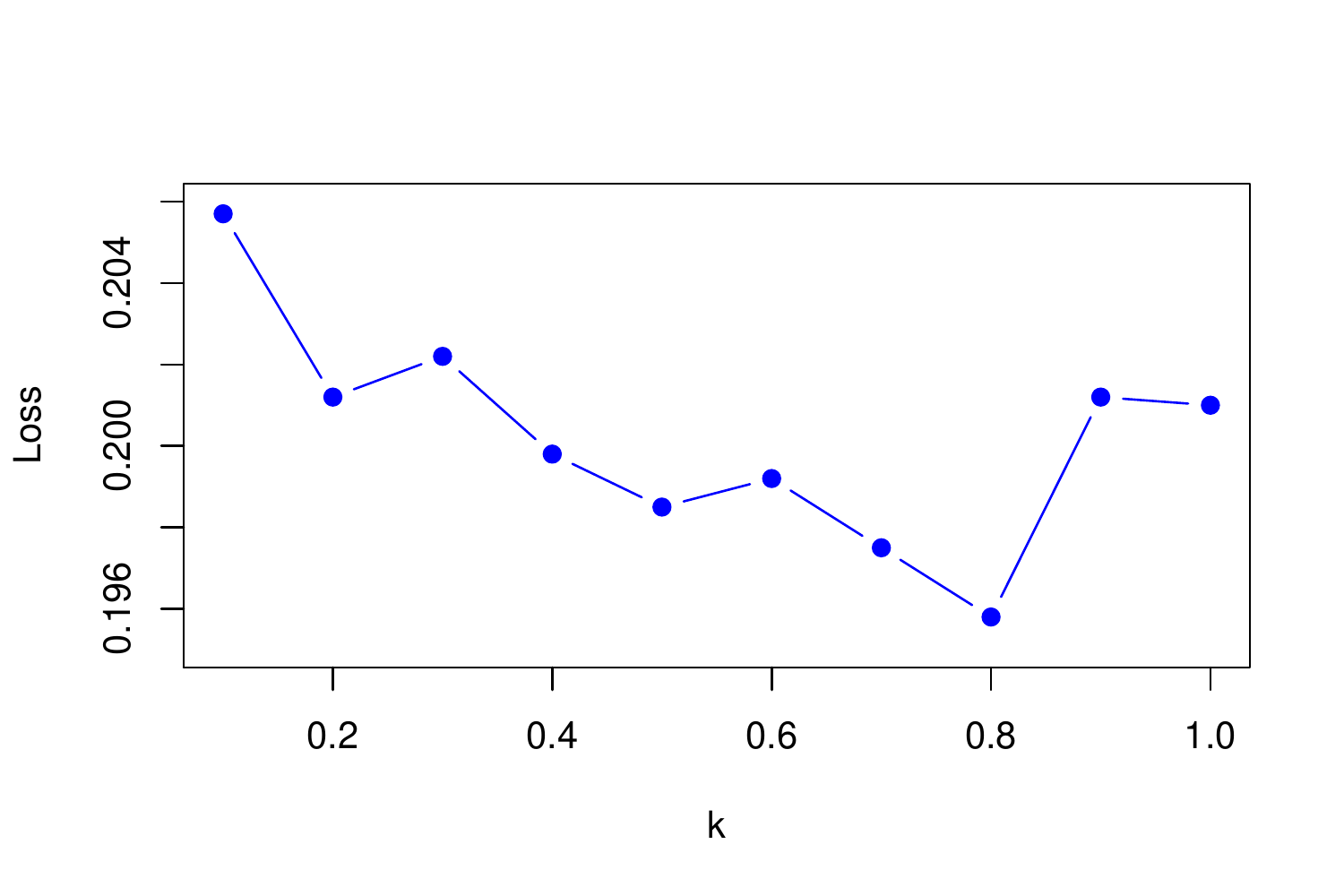}
\caption{The evolution of validation loss over decay speed $k$ for the exponential decay.}
\label{Decay_CA}
\end{figure}



\begin{table}[!h]
        \centering
        \caption{Performance comparison on KKBox for attrition prediction.}
        \label{KKBox_Metrics}
        \resizebox{\columnwidth}{!}{%
        \begin{tabular}{ccccc}
        \hline
        \multicolumn{1}{c|}{\textbf{Method}} & \textbf{AUC@ROC} & \textbf{MCC} & \textbf{F1 Score} & \textbf{AUC@PR}  \\ \hline

\multicolumn{1}{c|}{LR}	&	 0.8292 &	0.6560	&	0.6986  	&	0.7360  	\\ 

\multicolumn{1}{c|}{DNN}	&	 0.9016	&	0.6005   	&	 0.6552 	&	0.5937   \\ 

\multicolumn{1}{c|}{RF}	&	0.9394  &	 0.6961 	&	0.7314	&	0.8085  	 \\ 

\multicolumn{1}{c|}{NB}	&	0.5061	&	0.0314  	&	0.2467  	&	0.5671    \\ 

\multicolumn{1}{c|}{SVM}	&	0.5960    	&	0.2111    	&	0.1091   	& 0.4972	     \\ 

\multicolumn{1}{c|}{CNN}	& 0.8963	    	&	 0.5409 &	 0.5974	&	0.5263     \\ 

\multicolumn{1}{c|}{LSTM}  &	0.9293    &	  0.6786 	&	0.7171    	&  	0.7779    \\ \hdashline


\multicolumn{1}{c|}{BLA}	&	\textbf{0.9600}	&	\textbf{0.7280}	&	\textbf{0.7621}	&	\textbf{0.8436}  \\ \hline 

\end{tabular}%
        }
\end{table}

\subsubsection{Ablation Analysis}\label{FEATURE}
To explore the potential explanation for BLA's performance, a series of ablation experiments 
are conducted to study the role of key components of BLA.
We focus on KKBox here due to the limited observation time span of MOOCs.

\begin{figure}[!h]
\centering
\includegraphics[height=0.4\columnwidth, width=0.9\columnwidth]{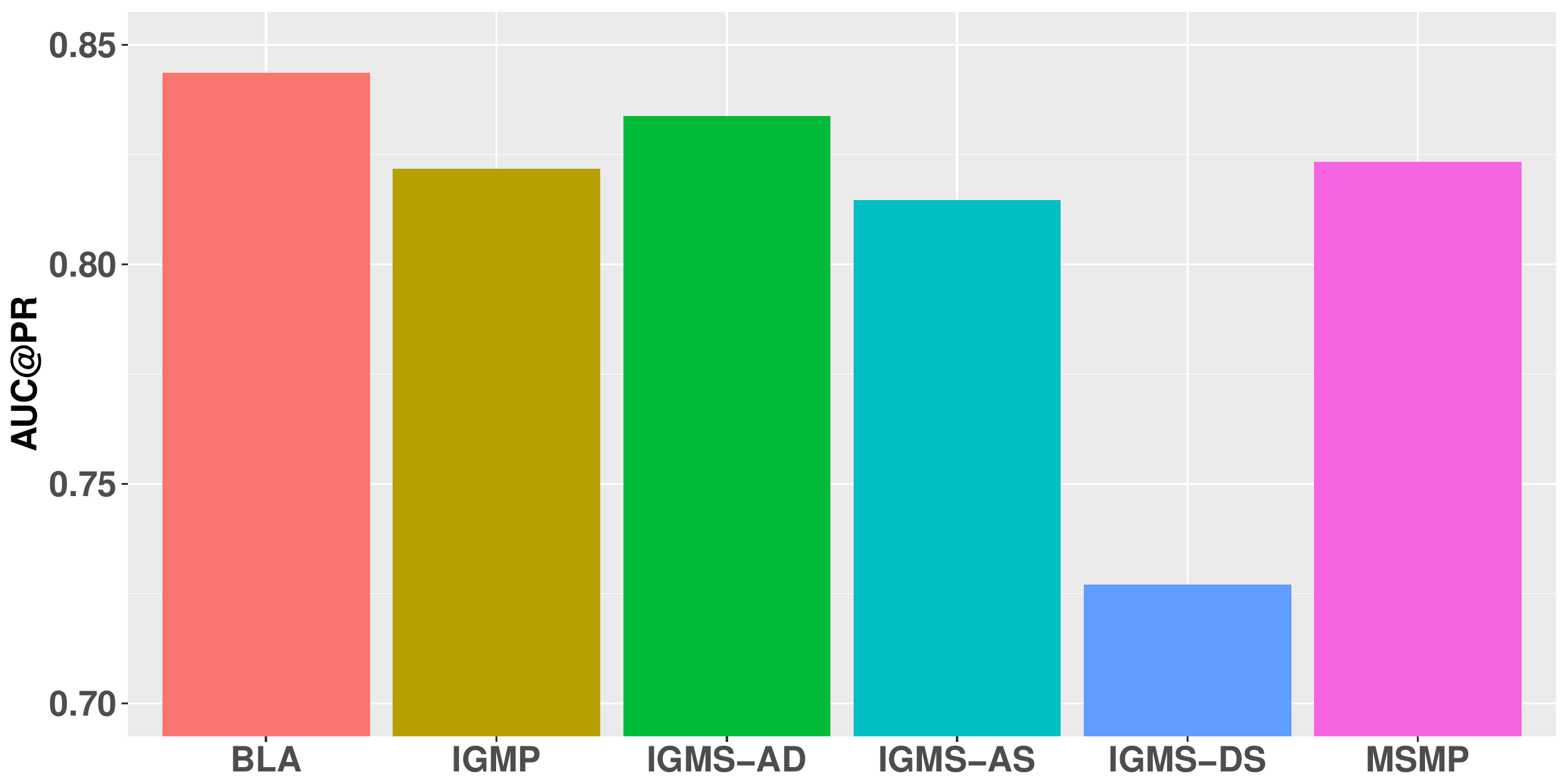}
\caption{Performance comparison for different variants of BLA.  Variants are described in Sec. \ref{Baselines}.}
\label{Component}
\end{figure}

First of all, we empirically study the decay mechanism.
As shown in Fig. \ref{decay}, $k = 1$ 
indicates all auxiliary statuses share equivalent weights in loss function, which also means that no decay mechanism is considered here.
Meanwhile, when $k \rightarrow 0$, it implies that auxiliary snapshot statuses are ignored and only the status of 
target time period is considered.  
The quasi U-shaped curve of loss on validation dataset demonstrates the existence of decay in attrition patterns over observed time steps 
as presented in Fig. \ref{Decay_CA}. 
Furthermore, the evidence that the value of right side is less than that of left side suggests 
the necessity of the proposed multi-snapshot strategy. 
The performance of different variants of BLA is also reported in 
Fig. \ref{Component}. 
Their performance disparity delivers useful points.
To be specific, 
it is activity path that has the highest impact, followed by dynamic path and finally static path.
Both intention guidance and multi-snapshot mechanisms shape BLA in different manners as well.

\begin{figure}[!h]
\centering
\includegraphics[height=0.5\columnwidth, width=\columnwidth]{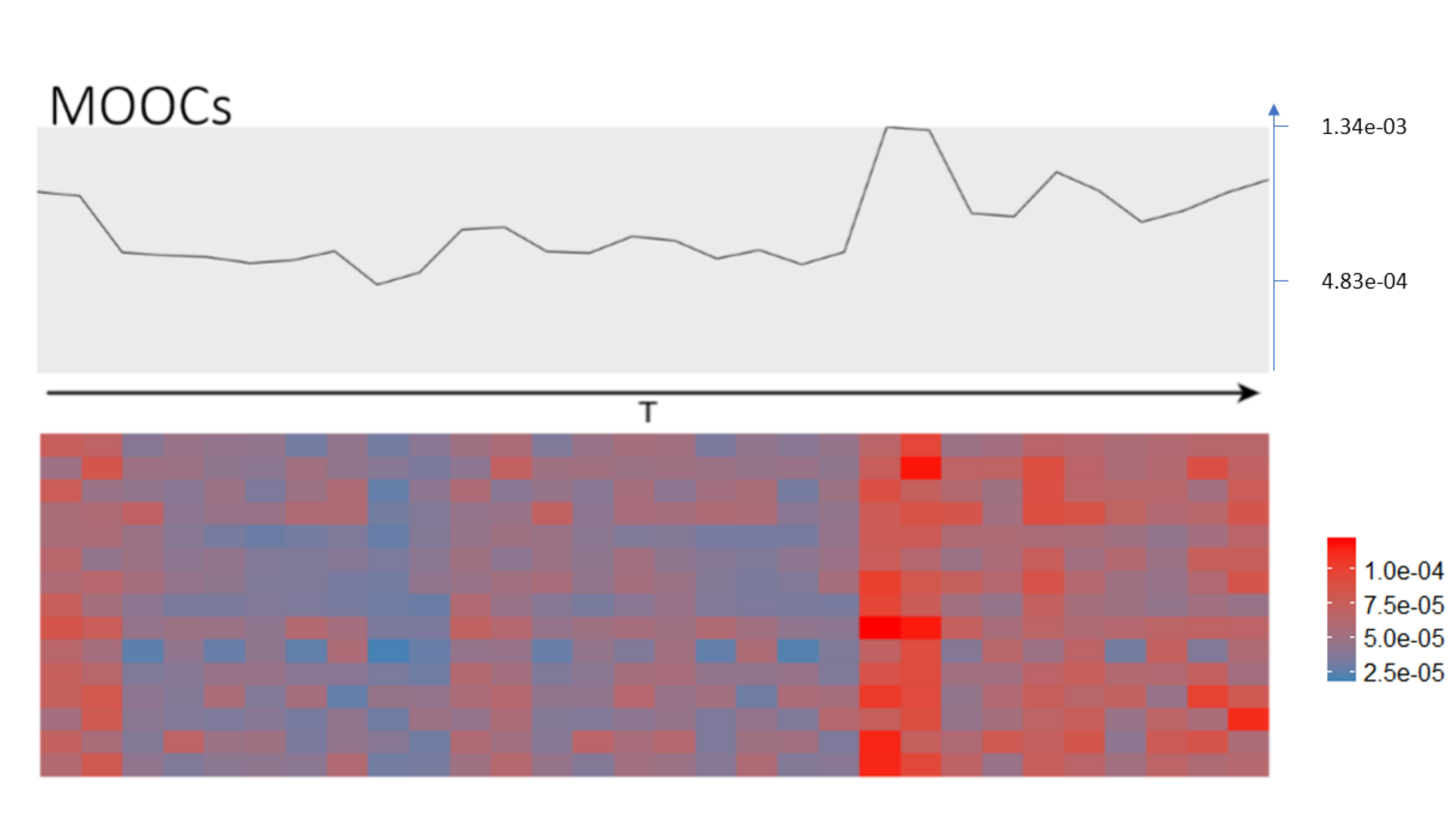}
\includegraphics[height=0.5\columnwidth, width=\columnwidth]{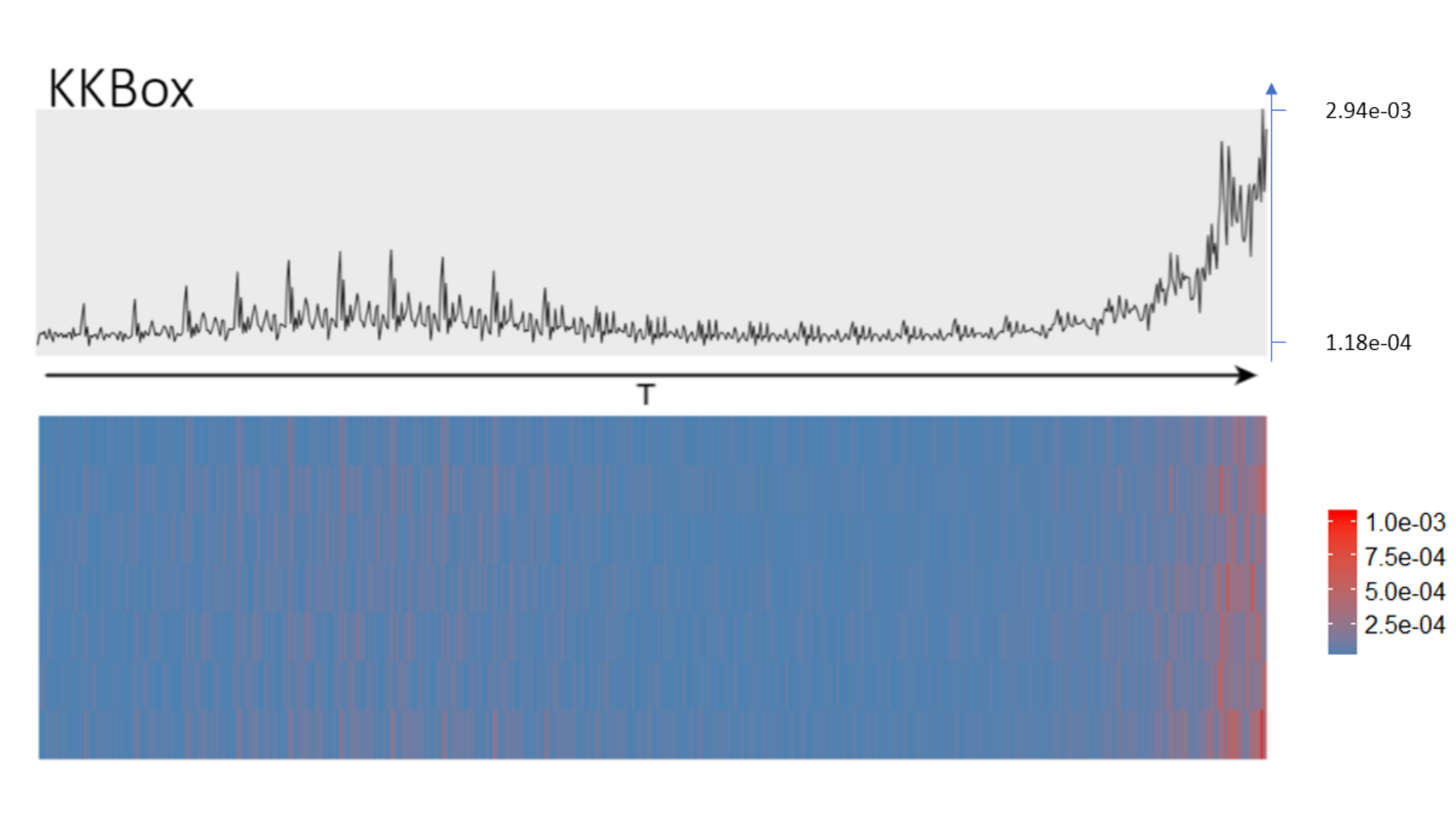}
\caption{Heat map of feature importance of the MOOCs and KKBox datasets.}
\label{MOOC_KKBox_Main}
\end{figure}

\subsubsection{Attrition and Retention Factors}
After attrition prediction, 
the next step typically is to identify underlying patterns/indicators 
or to explore feature importance. 

Regarding user activity logs, the feature importance across different observed time steps 
are visualized in Fig. \ref{MOOC_KKBox_Main}. 
Overall, the feature importance changes periodically with the peak value in the vicinity of the intersection of two successive snapshots. 
Furthermore, the peak increases roughly as observation moves onwards to the target time period. 
The locality of peak values indicates user activities around payments are informative and important compared with other time steps.
The evolution of peak values across different time periods shows that attrition 
within target time period is highly related to the proximate user activities, which is also intuitively reasonable. 

\begin{figure}[!h]
\centering
\includegraphics[height=0.2\columnwidth, width=0.9\columnwidth]{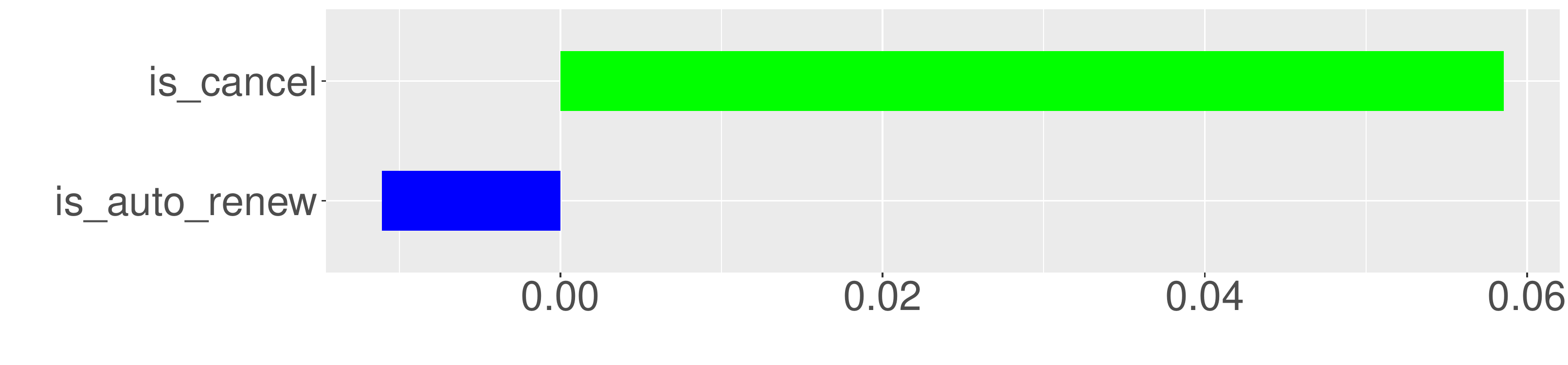}
\includegraphics[height=0.4\columnwidth, width=0.9\columnwidth]{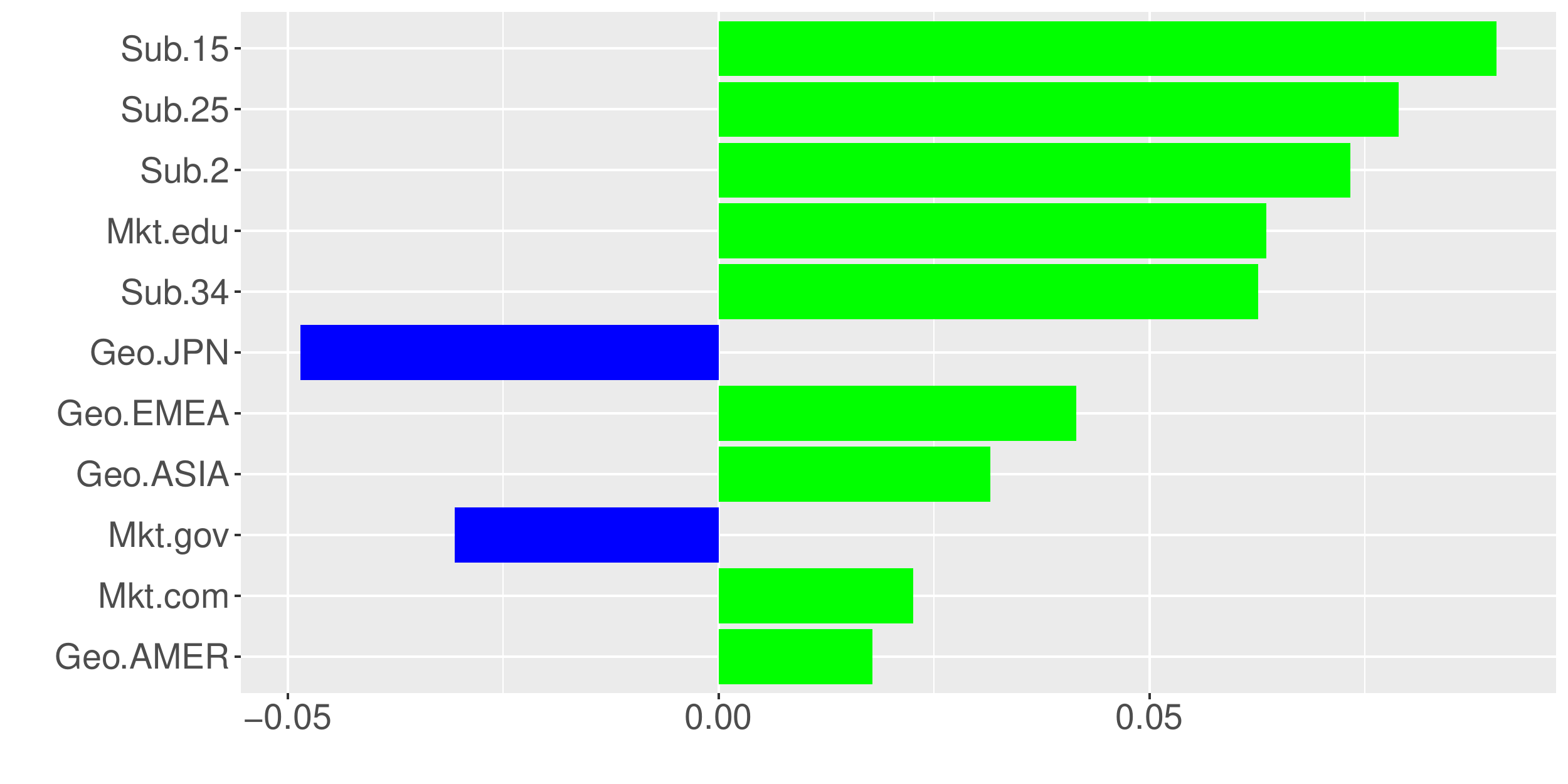}
\caption{Attrition and retention factors for KKBox (top) and Adobe CC (bottom). Features are detailed in Table \ref{Adobe_Profile}.}
\label{KKBox_Adobe_Aux}
\end{figure}

When it comes to dynamic and static user information, 
we also do in-depth analysis on KKBox. 
Among them, the most important features are  {\it is\_cancel}, and {\it is\_auto\_renew} from the dynamic side, and {\it registered\_via} from the static side.
As the registered method is provided anonymously, we cannot do any explanation.  
As shown in the top of Fig. \ref{KKBox_Adobe_Aux}, field {\it is\_cancel} indicates whether a user actively cancels a subscription or not,
which is proven to be positively correlated with attrition. It might be due to the change of service plans or other reasons, though. 
Naturally, feature {\it is\_auto\_renew} shows the intention of users to persist, which is also confirmed by the negative saliency value. 


\subsubsection{Adobe Creative Cloud}

\begin{table}[!b]
\caption{Basic statistics of Adobe CC users as of the beginning of the target time period. Observation span T (day), snapshot window size $\tau$ (day).}
\label{Adobe_Statistics}
\centering
\resizebox{0.5\textwidth}{!}{%
\begin{tabular}{c|c|c|c|c}
\hline
sampling & $\frac{\# \mathrm{\ of \ attrition}}{\# \mathrm{\ of \ persistence}}$ & T & $\tau$ &  target time period \\
\hline
 yes 	&     1    &   360   & 30 &  05/01/2017 $\sim$ 05/31/2017    \\ 
\hline
  yes	&     1    &   360  &  30 & 06/01/2017 $\sim$ 06/30/2017    \\ 	
\hline
  no	&     ---    &     360 & 30 &  07/01/2017 $\sim$ 07/31/2017    \\ 
\hline
\end{tabular}
}
\end{table}

\begin{table}
\centering
\caption{Activity logs of applications, dynamic and static information for Adobe CC users.}
\resizebox{0.5\textwidth}{!}{%
\begin{tabular}{ c|c|c }
\hline
 & Feature  & Remarks \\ \hline
\multirow{6}{*}{Activity}  
& Ps    &  booting times and total session time of  Photoshop \\ 
& Ai     &   booting times and total session time of Illustrator  \\ 
& Id    &   booting times and total session time of InDesign   \\
& Pr   &  booting times and total session time of  PremierePro   \\ 
& Lr    &  booting times and total session time of Lightroom     \\ 
& Ae     & booting times and total session time of  AfterEffects \\ 
& En     &  booting times and total session time of  MediaEncoder \\ 
\hline
\multirow{1}{*}{Dynamic} 
& Sub   &  the subscription age of Adobe CC \\ 
\hline
\multirow{2}{*}{Static} 
& Mkt    &  market segment (education, government, and commercial) \\ 
& Geo     &   general geographical code (JPN, EMEA, ASIA, AMER) \\ 
\hline
\end{tabular}
}
\label{Adobe_Profile}
\end{table}

Adobe CC provides entire collection of desktop and mobile applications for the brilliant design,
which is characterized by low user attrition rate. 
We apply the preliminary version of BLA (without decay mechanism or guided intention) called {\it pBLA}\footnote{Decay mechanism was not considered into our model during the internship period yet.} to perform 
churn prediction and analysis on sampled users, which are briefed in Table \ref{Adobe_Statistics}.
Concretely, user activity, dynamic and static information 
used in our model are described in Table \ref{Adobe_Profile}. 
Regarding activity logs, two daily metrics booting times and total session time
for each application (e.g., Photoshop) are recorded. 
Besides, we conduct both monthly and annual discretization 
of subscription age to capture two representative subscription types adopted by Adobe CC. 

\begin{figure}[!h]
\centering
\includegraphics[height=0.6\columnwidth, width=\columnwidth]{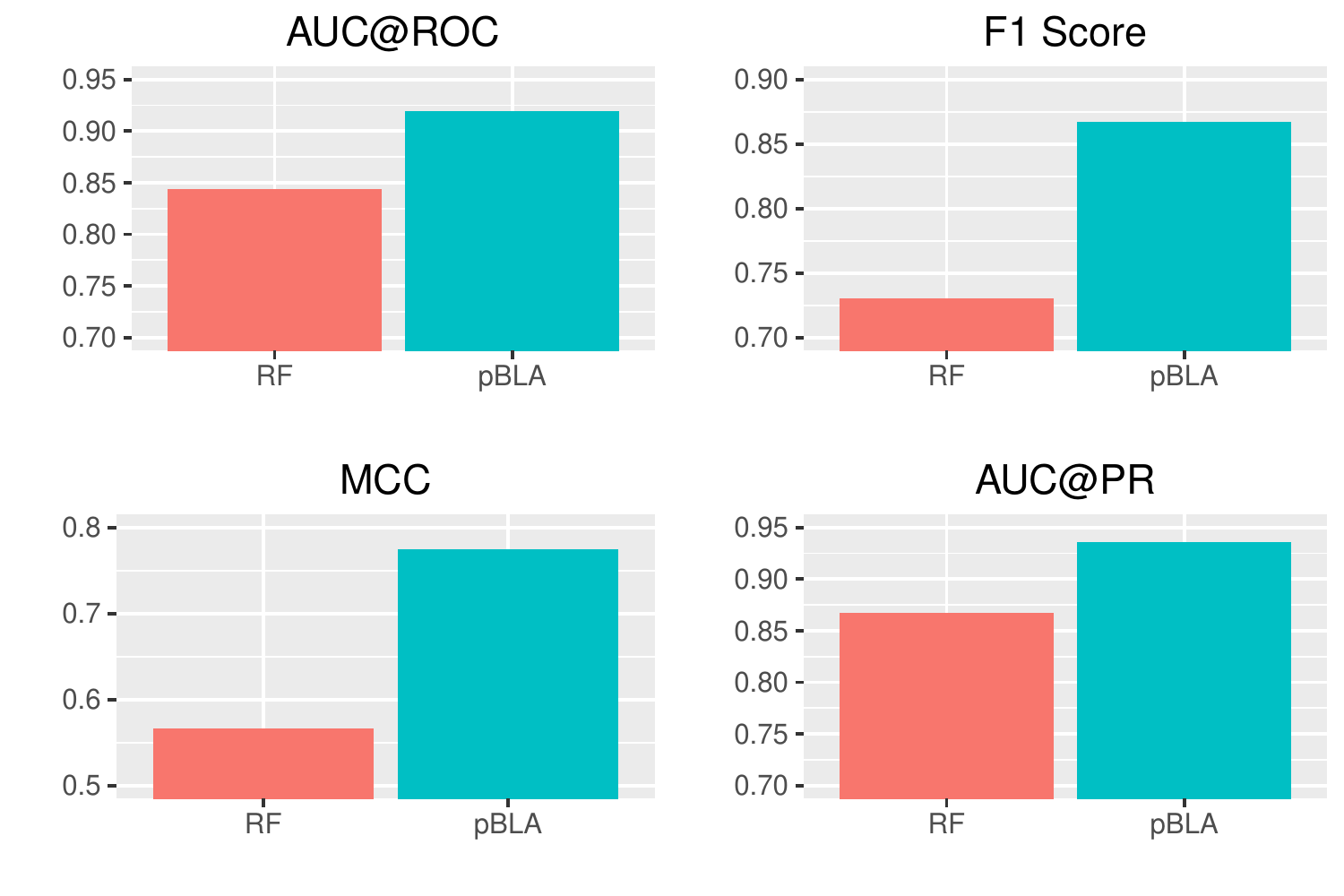}
\caption{Performance comparisons between pBLA and random forest.}
\label{Adobe_C1}
\end{figure}
\begin{figure}[!h]
\centering
\includegraphics[height=0.3\columnwidth, width=\columnwidth]{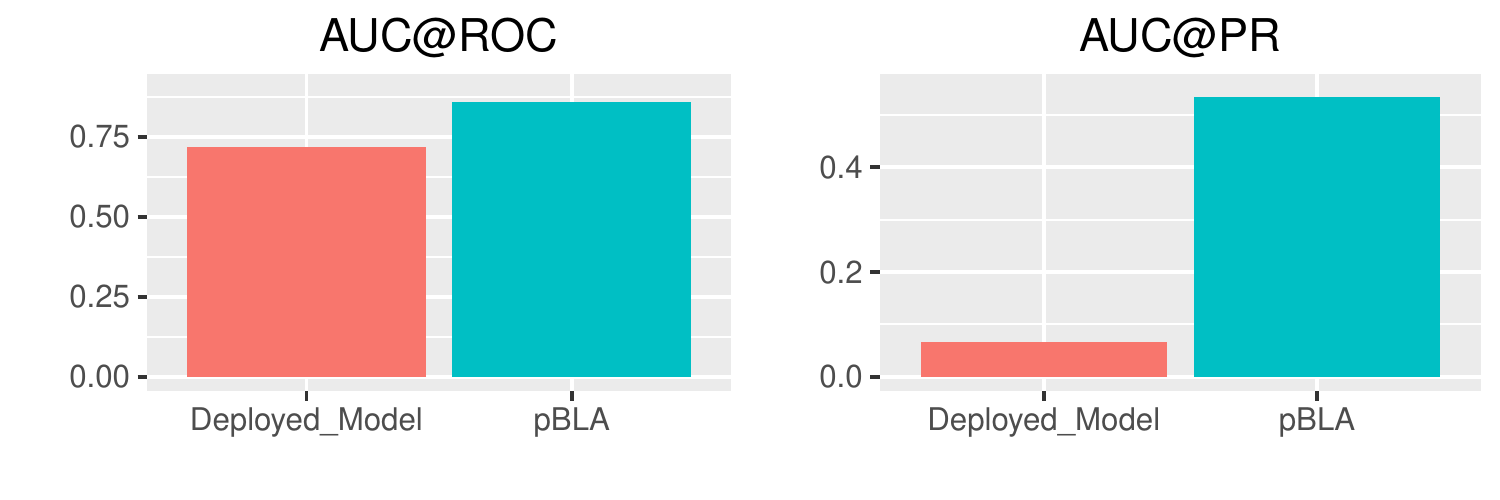}
\caption{Performance comparisons of pBLA against the currently deployed model.}
\label{Adobe_C2}
\end{figure}

In our experiments, we consider users with subscription age of  within 3 years. 
Due to confidentiality restrictions, we cannot disclose the volume of attrition and retention users. 
The dataset with the target time period of May 2017 is used for model development in which churned, and persistent users are sampled equivalently.
We then evaluate the predictive capacity of our algorithm in two scenarios. 
In the first scenario, the test dataset includes sampled users with a ratio of 1:1 during target time period of June 1 to June 30, 2017. 
We then compare pBLA with widely used random forest in the industry (e.g., Framed Data) for attrition prediction \cite{coussement2008churn,xie2009customer,spanoudes2017deep}. 
The results are reported in Fig. \ref{Adobe_C1}. The significant performance gain can be gained here.
In the other scenario, we compare our model with currently deployed model\footnote{Features are created based on user profile and products usage logs. 
For the product usage feature, we mainly utilized user usage records of 7 top Adobe CC products to 
generate counts, rates, recency over different time windows for different types of events. 
We also performed extensive feature engineering, such as imputation, capping, logarithm, binning, interactions of 
two variables like ratios and products.  Logistic regression based on multi-snapshot data
was trained with elastic net regularization. 
Model hyper-parameters are tuned based on 5-fold cross-validation with the best of efforts.} 
on users who were still active at the end of June 2017 (without sampling).
Our proposed model beats the currently deployed model greatly 
as reported in Fig. \ref{Adobe_C2}\footnote{The attrition probability adjustment of the currently deployed model is based on
all users beyond the subscribed age of 3 years. We thus omit threshold based evaluation metrics.}. 
The superiority of our algorithm over other approaches is more evident in Adobe CC than that in other datasets.
This mainly results from the difference of the subscription plans. 
Most subscriptions of Adobe CC are the type of annual plan while 
other datasets experience a couple of months (e.g., 30 to 90 days for most KKBox subscriptions).
The evolution of intended actions across long subscription plan period is amenable to our algorithm.

\begin{figure}[!h]
\centering
\includegraphics[height=0.48\columnwidth, width=\columnwidth]{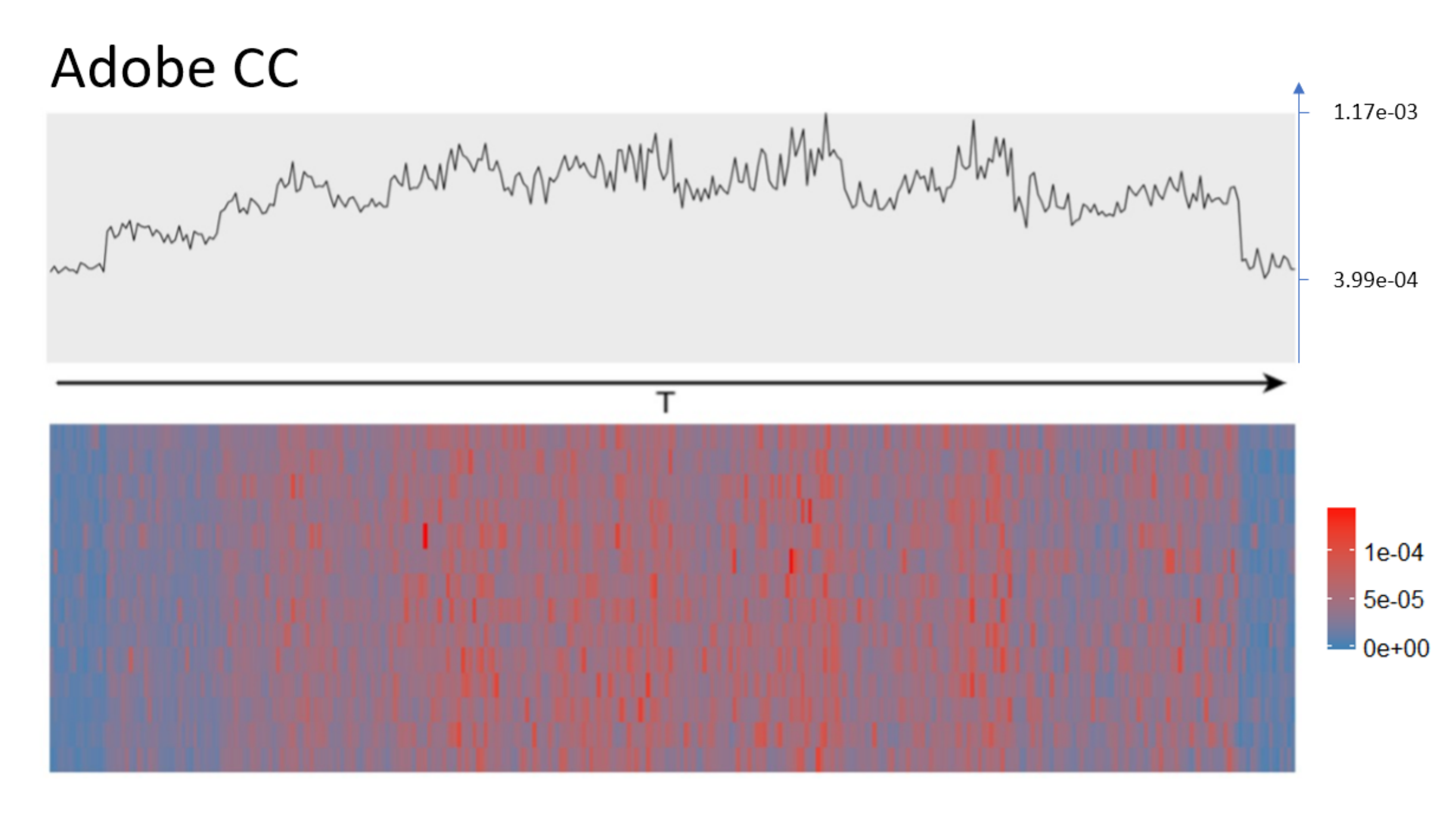}
\caption{Heat map of feature importance for Adobe CC user churn.}
\label{Adobe_Main}
\end{figure}

Likewise, the feature analysis implies that activity logs of users on applications of Adobe CC
are characterized by the explicit periodicity in terms of impacts on attrition as shown in Fig. \ref{Adobe_Main}.
Due to the long subscription plan for Adobe CC as mentioned before, the maximum of periodical peak values might be earlier than within the last month. 
Additionally, as shown in the bottom of Fig. \ref{KKBox_Adobe_Aux},  subscription age plays a very import role, for example, $15^{th}$, $25^{th}$, $2^{nd}$, $34^{th}$ month 
are the most risk months since the beginning of subscription,
which are all around the renewal dates of annual subscription plan\footnote{Monthly installment payment is available for the annual membership of Adobe CC.}. 
Regarding static information, Japan (JPN) is found to be the most persistent area compared with other geographical areas.
Also, it is easy to expect churn in subscribed users for the educational purpose, followed by the commercial and finally governmental purposes.

\section{Discussion and Future Work}\label{Discussion}
The introduced user alignment based on the calendar timeline enables an unbiased modeling.
The multi-path learning helps to fuse multi-view heterogeneous features, 
and the summarization layer is introduced to aggregate 
and integrate primitive user activity logs. In addition, we leverage
IGMS with decay mechanism to track evolving intentions. 
Finally, saliency maps are introduced to elucidate the activity patterns, attrition and retention factors.
There are some interesting aspects to explore in the future. 
First of all, from the perspective of the marketing campaign in the industry, 
the cost of attrition and retention may not be equivalent under some commercial circumstances. 
Thus, the probability threshold and corresponding loss function can be adaptively adjusted to 
account for their business profitability. 
In this case, some profit-driven strategies can be designed accordingly. 
Second, we consider the commonly used exponential decay in a trial-and-see way to 
explore the impacts of different time periods on the status of current time steps of interest.
The hyper-parameter $k$ is determined by the validation dataset. 
It is desirable to develop a principled and feasible way to tune $k$ automatically and even 
discover the underlying decay evolution involved
in the attrition prediction without the distribution assumption. 
This remains the topic of our future research. 


\section{Conclusion}\label{Conclusion}
In this work, we explore the classical attrition prediction (dropout and churn) problem and elucidate the underlying
patterns.  The proposed BLA is able to address an array of inherent
difficulties involved in traditional attrition prediction algorithms.
Particularly, the exploration of the decay mechanism further 
demonstrates the power and flexibility of our BLA in terms of
capturing the evolving intended actions of users.
The extensive experiments are conducted on two public real-world attrition datasets and  Adobe Creative cloud user dataset. 
The corresponding results show that our model can deliver the best performance over alternative methods with high feasibility.
The feature analysis pipeline also provides useful insights into attrition. 
Our work can also be applied to the attrition problem in related areas and other user intended actions. 

\section*{Acknowledgment}
We thank Sagar Patil for proofreading the manuscript.

\bibliographystyle{IEEEtran}
\bibliography{CustomerBehavior}

\end{document}